# AaP-ReID: Improved Attention-Aware Person Re-identification

Vipin Gautam, Shitala Prasad, and Sharad Sinha
School of Mathematics and Computer Science
Indian Institute of Technology Goa
{vipin2113106, shitala, sharad}@iitgoa.ac.in

## Abstract

*Person re-identification (ReID) is a well-known problem in the field of computer vision. The primary objective is to identify a specific individual within a gallery of images. However, this task is challenging due to various factors, such as pose variations, illumination changes, obstructions, and the presence of confusing backgrounds. Existing ReID methods often fail to capture discriminative features (e.g., head, shoes, backpacks) and instead capture irrelevant features when the target is occluded. Motivated by the success of part-based and attention-based ReID methods, we improve AlignedReID++ and present AaP-ReID, a more effective method for person ReID that incorporates channel-wise attention into a ResNet-based architecture. Our method incorporates the Channel-Wise Attention Bottleneck (CWA-bottleneck) block and can learn discriminating features by dynamically adjusting the importance of each channel in the feature maps. We evaluated Aap-ReID on three benchmark datasets: Market-1501, DukeMTMC-reID, and CUHK03. When compared with state-of-the-art person ReID methods, we achieve competitive results with rank-1 accuracies of 95.6% on Market-1501, 90.6% on DukeMTMC-reID, and 82.4% on CUHK03.*

## 1. Introduction

Person ReID stands as a computer vision task that entails the identification and correlation of individuals across disparate surveillance cameras. The core objective of person ReID is to ascertain whether an individual captured in one camera's viewpoint (referred to as a query image) corresponds to the same person observed in another camera's viewpoint (referred to as a gallery image). This task has gained substantial attention due to its vital role in diverse intelligent applications and video surveillance systems [3]. Within this landscape, Deep Learning (DL) based approaches have achieved remarkable advancements and exhibit superiority over their counterparts [35]. There are several applications and use cases where DL is including

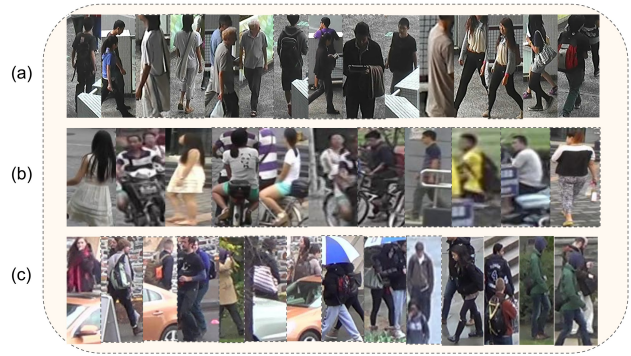

Figure 1. Challenges in person ReID: This figure presents examples of images collected from three different widely used datasets for person ReID. Each row corresponds to a dataset: (a) CHUK03, (b) Market1501, and (c) DukeMTMCReID, showcasing various examples of challenging training instances for person ReID.

hard and soft biometrics [19, 18, 17].

A majority of efforts have focused on global features, which are acquired through classical convolutional neural networks (CNNs) trained using classification loss and deep metric loss [9]. However, these methodologies often encounter challenges in scenarios marked by occlusions, obstacles, viewpoint variations, and changes in angle, as depicted in Fig. 1, rendering the problem even more intricate. To address these hurdles and enhance the learning process of CNNs, a variety of strategies have been proposed, encompassing part-based learning and attention-based methods.

Part-based learning involves segmenting person images into distinct parts, enabling the model to concentrate on specific clothing elements or body segments pivotal for identification. This approach effectively addresses challenges arising from pose variations and occlusions. Conversely, attention mechanisms empower the model to dynamically allocate focus to pertinent regions within an image, thereby capturing intricate details and contextual cues that substantially contribute to distinguishing individuals with similar appearances.

In this research, we introduce AaP-ReID, an extension of AlignedReID++[12]. Our framework introduces a CWA-bottleneck block aimed at extracting distinctive features like head, shoes, and backpacks from pedestrian images. This attention mechanism assigns varying weights to local features based on their relevance to the person ReID task. Consequently, it prioritizes the most discriminative features while disregarding less significant ones, leading to heightened discriminative power within the model. Through our experimental results, we demonstrate that part-based models can significantly benefit from integration with attention mechanisms. In summary, we make the following contributions.

- We introduce the CWA-bottleneck block for ResNet [5] to embed channel-wise attention, showcasing its efficacy through comprehensive experimentation.

- By substituting the bottleneck blocks of ResNet with the CWA-bottleneck block within the last two layers, we further enhance the discerning nature of the extracted features.

- We also analyze the importance of stride of the final downsample layer to effectively retaining more spatial information within the model's extracted features.

- The inclusion of batch normalization and dropout (BaND) on global feature maps serves to regularize the model and counteract overfitting tendencies.

The subsequent sections of this paper are structured as follows: Section 2 provides an overview of related research in the field of person ReID. Section 3 introduces the AaP-ReID architecture. Section 4 outlines the experimental configuration and the datasets employed in this study. Section 5 showcases the outcomes attained across three publicly accessible datasets. Lastly, we conclude in Section 6.

## 2. Related Work

Within the area of person ReID, two primary approaches emerge: representation learning and metric learning. Representation learning strategies center on acquiring a pedestrian representation that remains unaffected by factors like pose variations, lighting situations, and occlusions. Commonly, this entails utilizing CNNs to extract features from pedestrian images. On the other hand, metric learning approaches concentrate on acquiring a metric that gauges the similarity between two images of pedestrians using a loss function.

### 2.1. Representation Learning

Representation learning [13] can be partitioned into two distinct categories: Global and Local representations. Global representations encapsulate the overall persona, capturing elements such as clothing style, body shape, and overarching attributes. These representations provide a holistic view of the individual and are useful for initial matching. In the context of the person, the ReID network usually learns the global representation by training on softmax loss, often regarded as ID loss.

Furthermore, attention mechanisms have gained considerable traction, allowing networks to focus on informative regions within the global representation for improved discriminative power. Several attention methods focused on channel and spatial dimensions [6, 15, 28, 27] have been proposed to enhance the global representation potency of networks. Ye et al. [4] introduced attention-aware generalized mean pooling as a strategy for refining image retrieval. Chen et al. [1] introduced an occlusion-aware attention network with multiple branches. Zhang et al. [31] put forth a dynamic part-attention (DPA) approach for person ReID, employing a dynamic attention mechanism to steer the network towards the most distinctive body parts.

On a divergent note, local features hone in on specific regions or segments of an individual's body, such as the head, torso, or limbs. These features excel at capturing unique patterns like clothing accessories, tattoos, or distinctive poses. Particularly effective in scenarios where global appearances might be similar among different individuals or scale variation among the same instances. To learn these refined features researchers explored local features and proposed various methods based on part-based learning and multi-branch methods. Zhang et al. [33] introduced a multi-branch method to address pose misalignment concerns. Multiple approaches [22, 2, 23] have emerged to acquire refined local representations, often entailing the division of the entire pedestrian image into multiple horizontal segments with corresponding feature embeddings. Yet, these approaches contend with misalignment issues. Luo, Hao, and Jiang [12] introduced the DMLI method, striving to align local parts using the shortest path distance.

### 2.2. Deep Metric Learning

Deep metric learning (DML) [2], conversely, centers on acquiring a distance metric between pairs of data points. This typically involves transforming images into feature vectors and determining their similarity by assessing the distances separating them. A pair of images depicting the same individual constitutes a positive pair, whereas images portraying distinct individuals form a negative pair. DML's principal objective lies in minimizing the distance between positive samples within the learned metric space while simultaneously maximizing the distance between negatives.

Wang et al. [30] introduced a logistic discriminant metric learning method for person ReID, leveraging both original data and auxiliary data during training. Li et al. [32] de-vised a triplet focal loss tailored for person ReID, capable of elevating the significance of challenging triplets while de-emphasizing simpler ones. Zheng et al. [21] put forth a global hard sample mining technique for person ReID, utilizing a ranking list network in tandem with a multiplet loss.

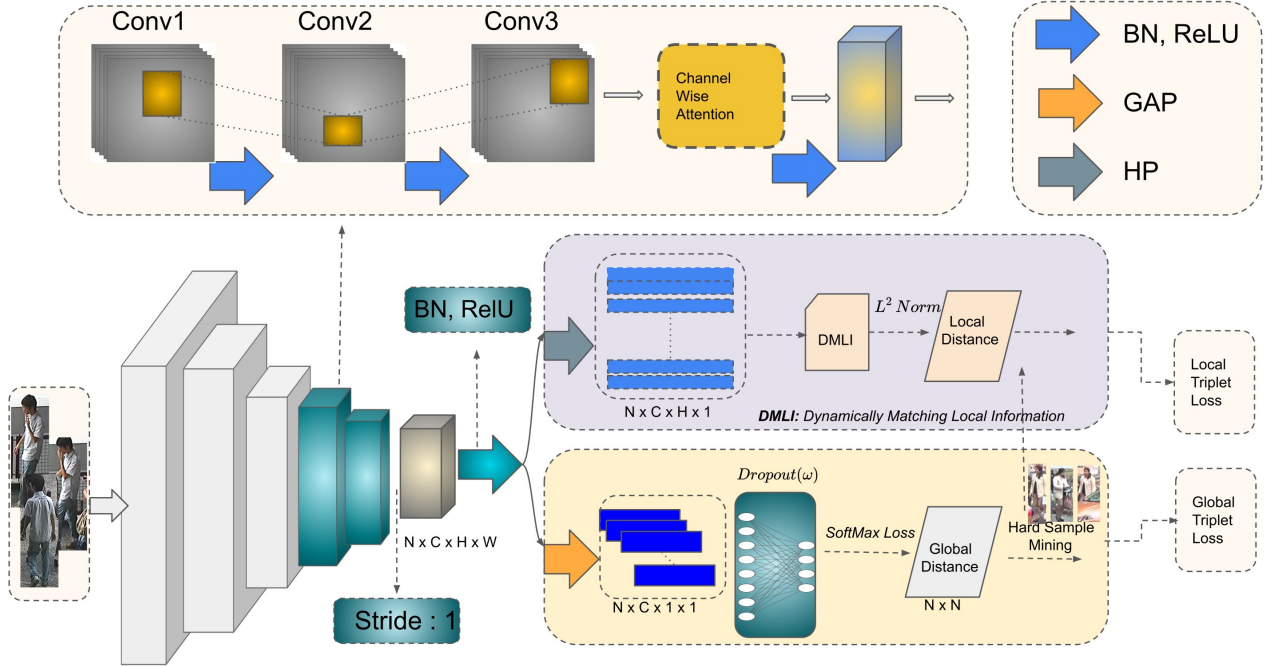

Figure 2. Overall design of the AaP-ReID. The blocks highlighted in green represent contributions to the baseline model, with the CWA-bottleneck block positioned at the top.

## 3. Attention-Aware Person ReID

Currently, the majority of person ReID algorithms rely on either part-based methods for image matching or attention mechanisms. Nonetheless, these techniques often falter in capturing pivotal features like the head, shoes, and backpacks, sometimes inadvertently emphasizing irrelevant attributes such as the torso—particularly evident when the subject is partially obscured. To tackle this limitation, we expanded upon a part-based approach and judiciously incorporated our novel CWA-bottleneck block. This section commences with an exposition of the overarching architecture of AaP-ReID, followed by an in-depth exploration of our proposed CWA-bottleneck block. Subsequently, we delve into the diverse training strategies implemented to elevate the performance of AaP-ReID.

### 3.1. Architecture of AaP-ReID

ResNet has consistently been favored for numerous person ReID algorithms, credited to its adeptness in extracting intricate features from images. We adopted ResNet as the foundational network for AaP-ReID, a choice facilitating comparisons with other state-of-the-art (SOTA) algorithms. The comprehensive network architecture is depicted in Fig. 2.

The backbone feature extractor generates a feature map $f$ with dimensions $N \times C \times H \times W$, where $N$ represents batch size, $C$ signifies channel count, and $H$ and $W$ stand for spatial dimensions. The ultimate convolutional layer's feature maps undergo processing via two distinct branches: the global and local branches. In the global branch, Global Average Pooling (GAP) is applied, culminating in a global feature vector of dimensions $N \times C \times 1$, followed by the computation of the $L_2$ norm. Ultimately, this global branch is subject to training through a fusion of ID and triplet losses.

While GAP effectively fails to capture spatial resemblances between images of individuals, the local branch assumes a pivotal role. The local branch endeavors to establish person-to-person similarities by aligning horizontal features through the employment of the DMLI [12] method. This local branch's horizontal pooling yields a feature vector with dimensions $N \times C \times H \times 1$. To quantify the local dissimilarity between two person images, a distance matrix $D$ is generated, subsequently employed to compute the total shortest path. This process culminates in the computation of a local triplet loss, leveraging hard samples identified via global distances within the global branch.

## 3.2. CWA-bottleneck Block

The CWA-bottleneck block is a simple but effective attention-based bottleneck that consists of three convolution blocks followed by a channel-wise attention (CWA) block, as shown in Fig. 2. The CWA block dynamically weighs the features according to their importance, allowing the CWA-bottleneck block to focus on the most important discriminative features and ignore the repeated and less important ones.

The CWA involves generating attention maps $f'$ for a given feature map $f$ with same shape $C \times H \times W$, where $C$ denotes the number of filters and $H$ and $W$ represent the spatial dimensions. Then they are subjected to global spatial pooling to create a channel descriptor which involved calculating the average value for each individual channels within that tensor, see the equation:

$$\mathcal{G}(f) = \frac{1}{H \times W} \sum_{i=1}^{H} \sum_{j=1}^{W} f_{i,j} \qquad (1)$$

where $\mathcal{G}()$ is a GAP operation performed to obtain $f'$ and $f_{i,j}$ is the value at $i$th row and $j$th column in $f$ feature map.

Next, the channel descriptor of shape $C \times 1 \times 1$ is passed through a Multi-Layer Perceptron (MLP) with the reduction parameter $r$, which helps to reduce the complexity of model and produces a descriptor of shape $\frac{C}{r} \times 1 \times 1$. The output is then processed through Batch Normalization (BN) [7] and a Rectified Linear Unit (ReLU) [14], see Eq. 2.

$$\mathcal{R}(f) = ReLU(BN(\mathcal{G}(f), r))) \qquad (2)$$

Then the attention weights are computed by passing the activated output through a second MLP using the same reduction parameter $r$ to adjust the dimensionality, as shown in Fig. 3. The resulting tensor, with shape $C \times 1 \times 1$, undergoes Sigmoid activation $\sigma$ and is then element-wise multiplied by $f$ to yield the attention feature maps as follows:

$$f' = f \odot \sigma(\text{MLP}(\mathcal{R}(f), r)) \qquad (3)$$

During the backpropagation stage of training, we compute the adjusted gradient for each channel $i$ as $g'_i = w_i \cdot g_i$, where $G = \{g_1, g_2, ..., g_c\}$ represents the gradient constituents corresponding to the channels while $w_i$ stands for the attention score assigned to the $i$th channel. This implies that the gradient contribution from channel $i$ is proportionally influenced by its attention score. By acquiring knowledge of these attention scores, the network highlights the significant channels while toning down the significance of the less crucial ones. It's plausible to think of the attention scores as coefficients that are applied to the gradients of each channel. As a result, the network gains the ability to magnify or diminish the impact of specific channels, based on their respective significance.

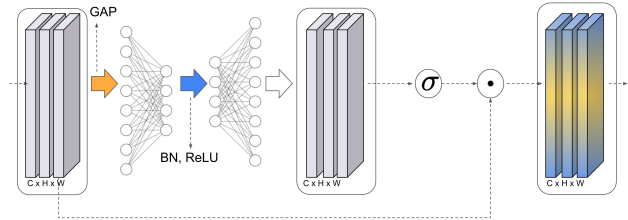

Figure 3. Structure of Channel Wise Attention block.

## 3.3. Impact of Strides

The convolutional operation's stride pertains to how many pixels the convolutional kernel shifts over the input image. When the stride is increased, the resultant feature map becomes smaller, causing information to be sacrificed. Conversely, a reduced stride generates a larger feature map, providing a greater amount of information for learning. To address this concern, we made an adjustment to the final downsampling step of the foundational network by altering the stride from 2 to 1. This alteration resulted in feature maps with dimensions of $16 \times 8$ instead of $8 \times 4$, thereby enhancing the scope of the area each unit covers and alleviating the loss of spatial information.

As a result, the model has gained the capacity to uphold intricate details, thereby enhancing its overall ability to distinguish. This alteration has enabled the model to maintain nuanced details that were previously not retained.

## 3.4. Integration of BN and Dropout (BaND)

BN and dropout are two well-known methods employed to enhance the performance of DL models. BN takes care of the normalization of activations in each layer, thereby promoting stable training processes and safeguarding against overfitting. Conversely, dropout involves the random omission of units in each layer, a strategy aimed at reducing the model's dependency on specific sets of features.

In our approach, we strategically implemented BN followed by *ReLU* activation right before the global branch, as shown in Fig. 2. The objective here was twofold: first, to acquire normalized feature maps denoted as $f_n$, and second, to eliminate non-linearities. This normalization step serves the purpose of steadying the features, consequently augmenting the model's capacity to generalize effectively. Ultimately, the GAP layer yields feature maps with dimensions $N \times C \times 1$.

For an added layer of resilience in the network, we introduced dropout subsequent to the GAP layer which acts as a regularization technique, countering overfitting by randomly nullifying a portion of activations during both forward and backward passes. Mathematically, this can be represented as follows:

$$\text{Dropout}(f_{\text{GAP}}) = f_{\text{GAP}} \odot Bernoulli(p) \qquad (4)$$

In this context, $f_{\text{GAP}}$ represents the outcome feature maps generated by the Global Average Pooling (GAP) layer. The symbol $\odot$ stands for element-wise multiplication, and $Bernoulli(p)$ refers to a binary dropout mask in which values are drawn from a Bernoulli distribution. Here, the dropout rate $p$ is set at $0.5$. By implementing dropout following GAP, we encourage the network to depend less on specific activations, thereby boosting its generalization capacity. This leads to a model that is more resilient and efficient.

The deliberate combination of BN, *ReLU* activation, and subsequent dropout in the final phases of the architecture collectively contribute to the overall improvement of the model's performance.

## 4. Experimental Settings

In this section, we commence by providing an overview of the datasets utilized in our experimental phase. Following that, we delve into the evaluation metrics chosen to assess the efficacy AaP-ReID. Lastly, we provide detailed insights into the implementation particulars of our approach, covering the nuances of the training process.

### 4.1. Datasets and Evaluation Metrics

We assessed the effectiveness of the suggested approach on three commonly employed datasets: Market-1501 [34], DukeMTMC-reID [20], and CUHK03 [10].

**Market1501** dataset encompasses 32,217 images capturing 1501 identified individuals, observed from six distinct camera viewpoints. It comprises a training subset with 751 unique identities and a testing subset with 750 distinct identities.

**DukeMTMCReID** comprises 36,411 images of 1812 individuals captured by eight cameras. The training subset of DukeMTMCReID includes 702 identities, while the testing subset encompasses 1110 identities.

**CUHK03** comprises 8765 images of 1467 labeled individuals. To ensure a fair comparison with the existing method, we adopted the same dataset division, where the training subset consists of 767 distinct IDs and the testing subset contains 700 distinct IDs.

To evaluate the performance of the proposed model, we employed two metrics: Cumulative Matching Characteristics (CMC) and Mean Average Precision (mAP). CMC gauges the likelihood of correctly identifying a pedestrian within the top k outcomes, while mAP quantifies the average precision across all results. Moreover, we applied the Re-Ranking technique [36] (RK). RK is a post-processing approach that enhances person ReID accuracy by refining the outcomes of the initial matching algorithm. To ensure fairness, we present results both with and without applying the RK technique.

### 4.2. Implementation Details

All experimentation was carried out employing the PyTorch framework [16]. AaP-ReID underwent optimization utilizing the ADAM optimizer, and the weight decay was configured to 0.0005. During the training phase, a batch size of 32 was employed, and the learning rate was set at 0.0002 with a step size of 150. The training process was executed over a total of 450 epochs. Images were resized to 256x128 for both training and testing. Data augmentation strategies, including random erasing and random horizontal flipping, were applied during training, along with normalization using mean values of (0.485, 0.456, and 0.406) and standard deviation values of (0.229, 0.224, and 0.225). Testing, on the other hand, only involved normalization.

## 5. Results and Discussion

In this section, we discuss the results of experiments conducted on the previously mentioned datasets. The examination provides a comprehensive exploration of both quantitative and qualitative discoveries. Initially, we delve into the outcomes derived from the ablation study. Subsequently, we compare AaP-ReID against currently prevailing state-of-the-art algorithms. Lastly, we present a qualitative analysis wherein we delve into salience maps and loss plots.

### 5.1. Ablation Study

In order to assess the efficacy of the introduced alterations to the baseline network, we conducted a comparative analysis of the performance enhancements resulting from the application of each modification. This encompassed the incorporation of the CWA-bottleneck block across all layers and specifically on the last two layers. Then, we conducted a comparative evaluation of the CWA-bottleneck block in contrast to various versions of attention mechanisms, such as the Bottleneck Attention Module (BAM) [15], Efficient Channel Attention (ECA) [27], Spatial Attention (SP) and Channel Wise Attention with Addition (CWA(+)).

**Impact of Training Strategies and CWA-bottleneck Block**

Tables 1 and 2 offer a comprehensive examination of how the CWA-bottleneck block and various adjustments introduced to the ResNet50 backbone at different layers influence performance. The first row illustrates the performance of the baseline model, AlignedReID++, without any alterations. The rows denoted as "Stride impact" and 'BaND' pertain to alterations in the stride of the final downsample layer of ResNet50 from 2 to 1 and the incorporation of a BN layer with dropout, respectively. 'CWA-all' signifies the configuration in which the CWA-bottleneck block replaces all the bottleneck blocks within ResNet50. Lastly, AaP-ReID demonstrate the outcomes when the CWA-bottleneck block substitutes the bottleneck blocks in the last two layers

Table 1. Impact of the training strategies and CWA-bottleneck block on ResNet50 backbone at different layers across different person ReID datasets

|  | Market1501 | | | DUKE-MTMC | | | CUHK03 | | |
|---|---|---|---|---|---|---|---|---|---|
| Method | maP | rank-1 | rank-5 | maP | rank-1 | rank-5 | maP | rank-1 | rank-5 |
| Baseline | 79.1 | 91.8 | 96.9 | 69.7 | 82.1 | 91.8 | 59.6 | 61.5 | 79.4 |
| Stride impact | 81.3(↑2.2) | 92.2(↑0.4) | 97.1(↑0.2) | 71.8(↑2.1) | 84.8(↑2.7) | 92.5(↑0.7) | 61.7(↑2.1) | 63.1(↑1.6) | 81.1(↑1.7) |
| BaND | 81.7(↑2.6) | 92.5(↑0.7) | 97.5(↑0.6) | 72.4(↑2.7) | 84.0(↑1.9) | 92.7(↑0.9) | 62.8(↑3.2) | 63.1(↑1.6) | 82.1(↑2.7) |
| CWA-all | 85.2(↑6.1) | 93.8(↑2.0) | 98.0(↑1.1) | 75.6(↑5.9) | 86.6(↑4.5) | 93.9(↑2.1) | 71.8(↑12.2) | 73.6(↑12.1) | 88.1(↑8.7) |
| *AaP-ReID* | 86.3(↑**7.2**) | 94.6(↑**2.8**) | 98.3(↑**1.4**) | 76.2(↑**6.5**) | 87.6(↑**5.5**) | 94.4(↑**2.6**) | 72.4(↑**12.8**) | 74.9(↑**13.4**) | 89.1(↑**9.7**) |
| Baseline* | 89.4 | 92.8 | 96.0 | 83.5 | 86.3 | 92.4 | 73.5 | 70.7 | 80.0 |
| Stride impact* | 91.0(↑1.6) | 93.4(↑0.6) | 96.3(↑0.3) | 86.2(↑2.7) | 88.6(↑2.3) | 93.1(↑0.7) | 76.0(↑2.5) | 72.8(↑2.1) | 82.0(↑2.0) |
| BaND* | 91.8(↑2.4) | 94.3(↑1.5) | 97.0(↑1.0) | 86.4(↑2.9) | 89.0(↑2.7) | 93.4(↑0.6) | 77.2(↑3.7) | 74.7(↑4.0) | 82.4(↑2.4) |
| CWA-all* | 93.3(↑3.9) | 94.9(↑2.1) | 97.5(↑1.5) | 88.3(↑4.8) | 89.9(↑3.6) | 94.2(↑1.8) | 84.2(↑10.7) | 82.1(↑11.4) | 89.4(↑9.4) |
| *AaP-ReID** | 93.9(↑**4.5**) | 95.6(↑**2.8**) | 97.7(↑**1.7**) | 88.6(↑**5.1**) | 90.6(↑**4.3**) | 94.8(↑**2.4**) | 84.7(↑**11.2**) | 82.4(↑**11.7**) | 89.5(↑**9.5**) |

**Bold**: superior, The symbols ↑' and *' indicate enhanced performance relative to the Baseline and Re-ranking (RK) algorithm results, respectively.

of ResNet50 (layers 3 and 4).

We progressively applied each technique to the baseline and assessed the resulting performance enhancement. Notably, substantial performance gains were observed after implementing 'BaND' and adjusting the 'stride'. For instance, there was an increase of 2.6% in mAP on Market1501, 2.7% on DUKE-MTMC, and 3.2% on CUHK03. Corresponding rank-1 accuracy scores saw improvements of 0.7%, 1.9%, and 1.6%, with rank-5 accuracy scores rising by 0.6%, 0.9%, and 2.7% for the respective datasets. Similarly, employing CWA-all yielded similar results, elevating mAP scores by 6.1%, 5.9%, and 12.2% on Market1501, DUKE-MTMC, and CUHK03, respectively. Hence, the integration of attention into the backbone architecture enhanced performance.

Through exhaustive experiments, we scrutinized the impact of the CWA-bottleneck block on specific layers. It was discerned that applying attention to the last two layers not only reduced the number of trainable parameters (as indicated in Table 2) but also significantly improved the performance. Specifically, we achieved an increase of 7.2%, 6.5%, and 12.8% in mAP scores, along with 2.8%, 5.5%, and 13.4% boosts in rank-1 accuracy scores on Market1501, DUKE-MTMC, and CUHK03, respectively.

In summary, by replacing the CWA-bottleneck block with the existing bottleneck block in the last two layers, our approach strikes a balance between performance and model complexity, positioning it as an optimal choice for person ReID. The proposed method not only attains superior accuracy results but also demonstrates a reduction in model parameters, rendering it a more efficient and effective solution for person ReID tasks.

**Attention Analysis**

Table 3 provides a comprehensive comparison of the impact of diverse attention techniques on different ResNet

Table 2. Comparison of (m) Parameter on ResNet50 Backbone at Different Layers Across Various Person ReID Datasets.

|  | **Market1501** | **DUKE-MTMC** | **CUHK03** |
|---|---|---|---|
| Baseline | 25.31 | 25.21 | 25.34 |
| CWA-all | 27.85(↑2.53) | 27.75(↑2.53) | 27.88(↑2.58) |
| *AaP-ReID* | 27.69(↑**2.37**) | 27.59(↑**2.37**) | 27.72(↑**2.42**) |

**Bold**: best. The Notation (m) indicates a million for parameters, hence least is desired.

backbones, using the Market1501 dataset. The considered attention methods include BAM, SP, ECA, CWA(+), and CWA(x). Notably, CWA(+) and CWA(x) represent distinct versions of the CWA-bottleneck block, wherein we explored varying feature fusion strategies—element-wise addition and multiplication, respectively. The BAM block integrates spatial and channel-wise attention by simultaneously computing spatial and channel attention in parallel branches, followed by an element-wise multiplication operation between the input feature maps and the consolidated attention maps from both branches. Thus, to dissect the effects of spatial and channel-wise attention, we conducted separate experiments for each.

The outcomes consistently indicate that the proposed approach, CWA(x) outperforms the other attention methods across a range of ResNet variants including ResNet18, ResNet34, ResNet50, and ResNet101. As an illustration, when considering ResNet50 with CWA(x), it attains rank-1 and rank-5 accuracies of 94.6% and 98.3%, respectively, alongside an mAP score of 86.3%. In contrast, BAM, despite having more parameters (30.20M), achieves a lower rank-1 accuracy of 94.3%, rank-5 accuracy of 98%, and slightly lower mAP score of 86.2%. Furthermore, the superiority of CWA(x) extends to ResNet18, ResNet34, and ResNet101 over BAM and SP. While these other variants utilize more parameters, CWA(x) manages to outperform them while employing fewer parameters.

Table 3. Comparison of attention methods impact on various ResNet backbones with the proposed Approach CWA(x) on Market1501 dataset

| Backbone | Metrics | BAM | SP | ECA | CWA(+) | CWA(x) |
|---|---|---|---|---|---|---|
| ResNet18 | maP | 74.5 | 70.7 | 73.60 | 75.70 | **75.80** |
| | rank-1 | 88.2 | 86.3 | 87.1 | 88.50 | **89.40** |
| | rank-5 | 95.5 | 96.6 | 94.9 | **97.50** | 96.20 |
| | params | 11.80 | 11.72 | 11.62 | 11.71 | 11.71 |
| ResNet34 | maP | 77.9 | 74.8 | 78.5 | 79.9 | **80.3** |
| | rank-1 | 89.5 | 88.3 | 90 | 90.6 | **91.5** |
| | rank-5 | 95.7 | 95.2 | 96.3 | 96.7 | **96.9** |
| | params | 22.04 | 21.89 | 21.73 | 21.88 | 21.88 |
| ResNet50 | maP | 86.2 | 84.4 | 85.3 | 85.6 | **86.3** |
| | rank-1 | 94.3 | 93.3 | 93.4 | 93.8 | **94.6** |
| | rank-5 | 98 | 97.7 | 97.7 | 98.2 | **98.3** |
| | params | 30.20 | 27.83 | 25.31 | 27.69 | 27.69 |
| ResNet101 | maP | 85.7 | 83.6 | 85.9 | 85.7 | **86.6** |
| | rank-1 | 93.4 | 92.5 | 93.9 | 93.8 | **94.2** |
| | rank-5 | 97.6 | 97.2 | **98.6** | 98.0 | 98.2 |
| | params | 53.82 | 49.20 | 44.30 | 48.93 | 48.93 |

## 5.2. Comparison with State-of-the-Art

Table 4 presents a summarized account of our thorough competitive analysis. We extensively compared our work with state-of-the-art person ReID methodologies, encompassing SNR [8], CBN [37], CAP [26], CtF [24], APR [11], HOReID [25], OAMN [1], SGAM [29], and AlignedReID++ [12]. Given that our work is built upon AlignedReID++, we employed it as the baseline. Furthermore, all the methods under consideration utilize ResNet as a feature extractor.

On Market-1501, we attained a mAP score of 86.3% and a notable rank-1 accuracy of 94.6%, the highest among all the methods. Within the context of DukeMTMC-reID, we outperformed all methods, achieving a mAP score of 76.2% and a rank-1 accuracy of 87.6%, which only slightly trails behind CAP. For CUHK03, we adopted detected bounding boxes as part of the testing protocol and achieved a commendable mAP score of 67.2% alongside a robust rank-1 accuracy of 81.4%. When compared to AlignedReID++ in conjunction with RK, AaP-ReID demonstrates substantial improvement, yielding a mAP of 93.9% and a rank-1 accuracy of 95.6%. This represents an enhancement of 4.5% and 2.8%, respectively. Similar performance gains are evident when contrasting our work with plain AlignedReID++, where we achieve a mAP of 84.7% and a rank-1 accuracy of 82.4%.

## 5.3. Qualitative Results

**Heatmaps visualization**

Table 4. SOTA Comparison on different person ReID Datasets

| Method | Market1501 | | DukeMTMC | | CUHK03 | |
|---|---|---|---|---|---|---|
| | maP | rank-1 | maP | rank-1 | maP | rank-1 |
| SNR[8] | 84.7 | 94.4 | - | - | - | - |
| CBN[37] | 77.3 | 91.3 | 67.3 | 82.5 | - | - |
| CAP[26] | 85.1 | 93.3 | 76.0 | **87.7** | - | - |
| CtF[24] | 84.9 | 93.7 | 74.8 | 87.6 | - | - |
| APR[11] | 66.8 | 87.04 | 55.6 | 73.9 | - | - |
| HOReID[25] | 84.9 | 94.2 | 75.6 | 86.9 | - | - |
| OAMN[1] | 79.8 | 93.2 | 72.6 | 86.3 | - | - |
| SGAM[29] | 77.6 | 91.4 | 67.3 | 83.5 | - | - |
| A-ReID++[12] | 79.1 | 91.8 | 69.7 | 82.1 | 59.6 | 61.5 |
| AaP-ReID | **86.3** | **94.6** | **76.2** | 87.6 | **72.4** | **74.9** |
| A-ReID++[*] | 89.4 | 92.8 | 83.5 | 86.3 | 73.5 | 70.7 |
| AaP-ReID[*] | **93.9** | **95.6** | **88.6** | **90.6** | **84.7** | **82.4** |

**Bold**: best, A-ReID++ is shortform for AlignedReID++, The superscript[*] represents RK is used.

Saliency maps serve as visual representations that highlight the significant areas within an image that influence the decision-making process of DL model. For the purpose of qualitative analysis, we generated saliency maps for both the AaP-ReID and baseline models. Fig. 5 offers a comparison of the heatmaps produced by these two models. The top row showcases the original pedestrian images, while the subsequent two rows exhibit the heatmaps generated by the AaP-ReID and baseline models, respectively.

Through a range of examples, we effectively illustrate the prowess of our approach in scenarios involving obstructed pedestrians, obstacles, or partially visible subjects. In row (b) of Fig. 5, AaP-ReID adeptly focuses on discriminative pedestrian attributes, capturing overarching global traits like 'face,' 'shoes,' and 'backpacks,' while attenuating less universal characteristics such as 'box' and 'bicycle.' Conversely, the existing method struggles to encapsulate these distinguishing attributes. For instance, when faced with occlusions, it remains fixated on the occluded areas instead of emphasizing features like the face or legs, which serve as distinguishing traits. This marked differentiation underscores the effectiveness of our proposed method in recognizing more generalized and distinctive global features.

**Loss Analysis**

To showcase the enhanced convergence achieved through the proposed CWA-bottleneck block, we present loss curves for ResNet18, ResNet34, ResNet50, and ResNet101 in Fig. 4. Each curve visualizes the progression of loss for ECA, CWA(x), SP, CWA(+), and BAM, specifically applied to layers 3 and 4 in our approach.

The depicted plots reveal that for smaller networks like ResNet18 and ResNet34, SP exhibited inadequate convergence, failing to effectively reach the global minimum across all methods. Conversely, most methodologies displayed convergence within the epoch range of 340 to 380. Notably, our approach, CWA(x), consistently demonstrated the lowest loss across all backbone variations when compared to other techniques. The efficacy of our method is underscored by its substantial convergence improvement.

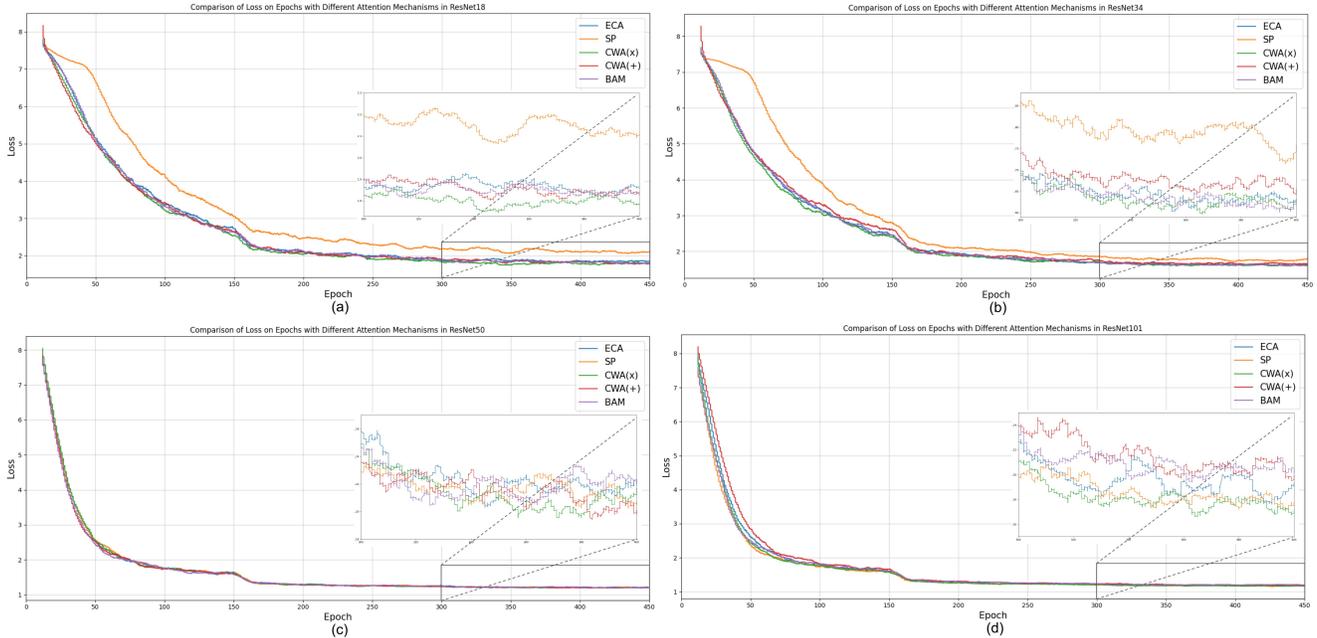

Figure 4. Training Loss vs Epoch plots for different attention methods on ResNet18, ResNet34, ResNet50, and ResNet101.

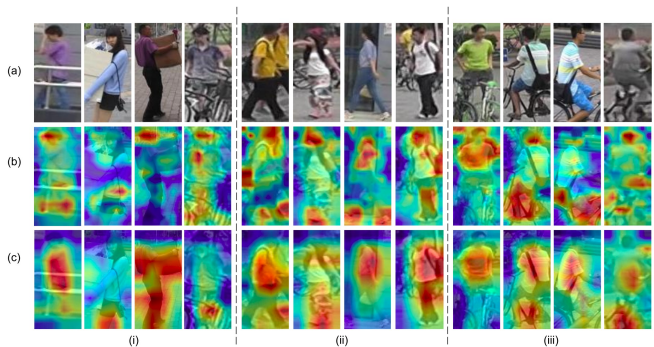

Figure 5. Saliency map visualization comparing our approach to the baseline method. (a) Original images, (b) Our heatmaps, and (c) Baseline heatmaps. Each block showcases four examples: (i) Pedestrians with obstacles/occlusion, (ii) General Pedestrians images, and (iii) Pedestrians on vehicles.

Upon a closer examination, the loss curves exhibit a gradual downward trend with minor fluctuations. This consistent pattern suggests that the model is progressively absorbing and adapting to the data, fostering sustained learning performance. Notably, our model maintains its dynamic performance, exhibiting similar proficiency when compared to other state-of-the-art backbone networks.

## 6. Conclusion

In this paper, we proposed a novel person ReID method that incorporates channel-wise attention into a ResNet-based architecture. The CWA-bottleneck block is able to learn discriminative features by dynamically adjusting the importance of each channel in the feature maps. We evaluated our method on three benchmark datasets: Market-1501, DukeMTMC-reID, and CUHK03. We achieved competitive results on all three datasets, with a rank-1 accuracy of 95.6% on Market-1501, 90.6% on DukeMTMC-reID, and 82.4% on CUHK03. These results demonstrate the effectiveness of our method on a variety of challenging datasets. Additionally, we conducted a systematic exploration of the CWA-bottleneck block, incorporating it into different ResNet backbones and benchmarking it against prominent attention techniques. Our results showed that the CWA-bottleneck block consistently outperforms other methods, demonstrating its superior efficacy.